\documentclass[10pt,twocolumn,letterpaper]{article}

\usepackage[pagenumbers]{cvpr} 

\usepackage{graphicx}
\usepackage{amsmath}
\usepackage{amssymb}
\usepackage{booktabs}
\usepackage{float}
 \usepackage{adjustbox}
\usepackage{multirow}
\usepackage{colortbl}

\definecolor{cvprblue}{rgb}{0.21,0.49,0.74}
\usepackage[pagebackref,breaklinks,colorlinks,allcolors=cvprblue]{hyperref}

\usepackage{algorithm}  
\usepackage{algorithmicx}
\usepackage{algpseudocode}
\usepackage[capitalize]{cleveref}
\crefname{section}{Sec.}{Secs.}
\Crefname{section}{Section}{Sections}
\Crefname{table}{Table}{Tables}
\crefname{table}{Tab.}{Tabs.}

\def\confName{CVPR}
\def\confYear{2025}

\title{EADReg: Probabilistic Correspondence Generation with Efficient Autoregressive Diffusion Model for Outdoor Point Cloud Registration}

\author{Linrui Gong\\
Shanghai Jiao \\
{\tt\small ttrr2021@sjtu.edu.cn}
\and
Jiuming Liu\\
Shanghai Jiao Tong University\\
{\tt\small liujiuming@sjtu.edu.cn}
\and
 Junyi Ma\\
Shanghai Jiao Tong University\\
{\tt\small junyi.ma@sjtu.edu.cn}
\and
 Lihao Liu\\
University of Cambridge\\
{\tt\small ll610@cam.ac.uk}
\and
 Yaonan Wang\\
Hunan University\\
{\tt\small yaonan@hnu.edu.cn}
\and
 Hesheng Wang\thanks{Corresponding Author}\\
Shanghai Jiao Tong University\\
{\tt\small wanghesheng@sjtu.edu.cn}
}

\begin{document}
\maketitle

\begin{abstract}
\label{abs}


Diffusion models have shown the great potential in the point cloud registration (PCR) task, especially for enhancing the robustness to challenging cases. However, existing diffusion-based PCR methods primarily focus on instance-level scenarios and struggle with outdoor LiDAR points, where the sparsity, irregularity, and huge point scale inherent in LiDAR points pose challenges to establishing dense global point-to-point correspondences. To address this issue, we propose a novel framework named EADReg for efficient and robust registration of LiDAR point clouds based on autoregressive diffusion models. EADReg follows a coarse-to-fine registration paradigm. In the coarse stage, we employ a Bi-directional Gaussian Mixture Model (BGMM) to reject outlier points and obtain purified point cloud pairs. BGMM establishes correspondences between the Gaussian Mixture Models (GMMs) from the source and target frames, enabling reliable coarse registration based on filtered features and geometric information. In the fine stage, we treat diffusion-based PCR as an autoregressive process to generate robust point correspondences, which are then iteratively refined on upper layers. Despite common criticisms of diffusion-based methods regarding inference speed, EADReg achieves runtime comparable to convolutional-based methods. Extensive experiments on the KITTI and NuScenes benchmark datasets highlight the state-of-the-art performance of our proposed method. Codes will be released upon publication.

\end{abstract}

\section{Introduction}
\label{sec:intro}

\begin{figure}
\centering
\includegraphics[scale=0.35]{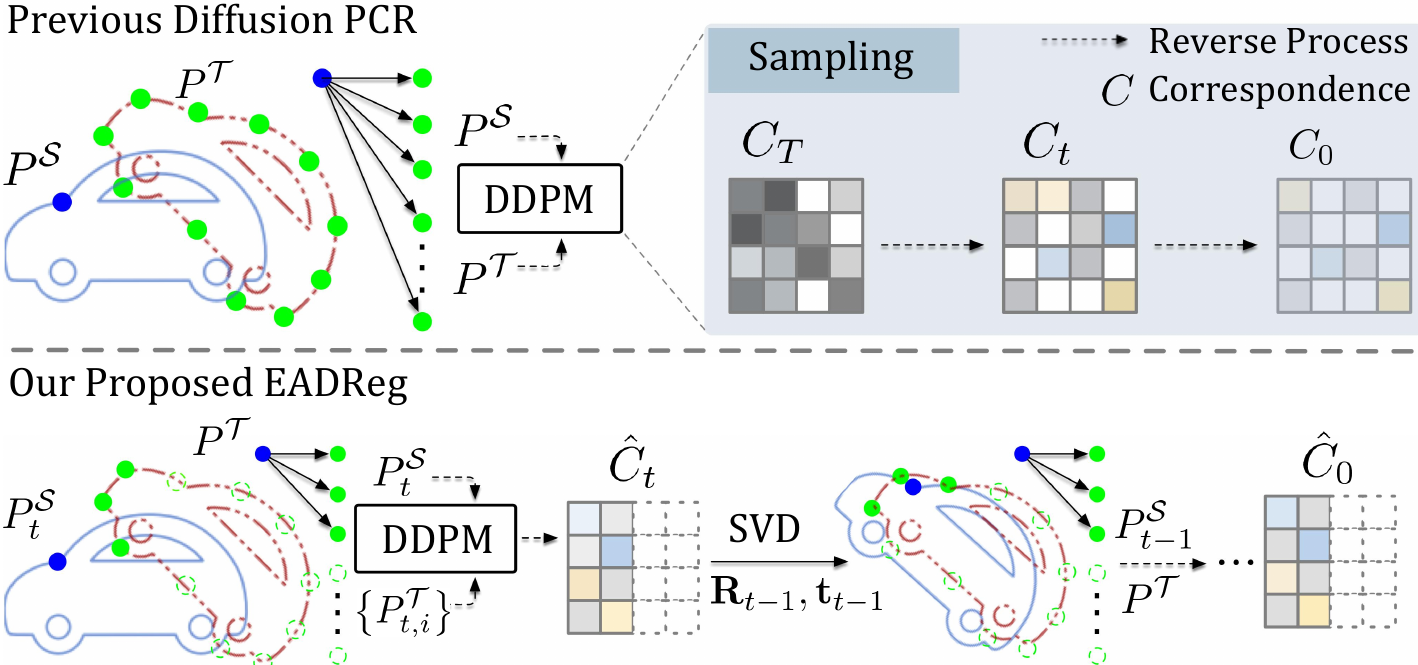}
\caption{\textbf{Comparison with previous Diffusion PCR methods.} Previous diffusion-based PCR methods directly generate global dense point-to-point (P2P) correspondences $C_{t}\in\mathbb{R}^{N^{\mathcal{S}}\times N^{\mathcal{T}}}$, resulting in high training costs. In contrast, our proposed EADReg predicts correspondences within the top-K nearest neighbors of the source points $\hat{C}\in\mathbb{R}^{N^{\mathcal{S}}\times K}$, leveraging reliable coarse registration. Besides, We integrate an autoregressive framework into the reverse process to better suit PCR tasks.}
\label{fig.loopdiff}
\end{figure}

\vspace{-3mm}

Point cloud registration (PCR) in outdoor environments is a fundamental task, which aims to compute the 6-DOF rigid transformation between two LiDAR frames. It is essential to various downstream applications, such as global localization~\cite{Elbaz_2017,Pomerleau_2015,Sun_2022,Xie_2021}, SLAM~\cite{Cattaneo_2022,Hitchcox_2020,Jiang_2023_nav}, navigation and auto-driving applications~\cite{liu2023translo,liu2024difflow3d,liu2025dvlo,Miao_2016,Kim_2019,deng2022trafficcam,Pomerleau_2015}. The formal procedure for PCR involves extracting reliable point features using hand-crafted or learning-based methods, introducing alignment algorithms to establish point-to-point (P2P) correspondences, and finally calculating the transformation parameters via singular value decomposition (SVD).

Establishing accurate P2P correspondences is a key issue in PCR~\cite{GeoTransformer,chen2023diffusionpcr,zou2023learning,liu2024mamba4d}. Recent methods have concentrated on enhancing the reliability of point correspondences. GeoTransformer~\cite{GeoTransformer} improves the correspondence establishment by introducing the geometric transformer into PCR tasks. DiffusionPCR~\cite{chen2023diffusionpcr} constructs a powerful transformer feature extraction module to ensure high-quality correspondence generation. RoReg~\cite{roreg} introduces oriented descriptors with both rotation equivariance and invariance to enhance correspondences. However, due to noisy inputs and unreliable feature correlations, previous regression-based methods still struggle to find accurate P2P correspondences 

Distinct from regression methods, we resort to diffusion models in this paper to improve the robustness for P2P estimation with a correspondence-generation manner. Recently, Denoising Diffusion Probabilistic Models (DDPMs) have demonstrated a strong ability in domains such as image and video generation, segmentation, image matching, and motion prediction~\cite{Zhang_2023,ho2022imagen,liu2023traffic,Jiang_2023,diffmatch}, which generate realistic predictions from standard Gaussian noise. Inspired by these advancements, Some studies attempt to extend diffusion models into point cloud registration~\cite{chen2023diffusionpcr,diffpcr,she2024pointdifformer,se3diff} for accurate dense correspondence generation and transformation matrix regression. DiffusionPCR~\cite{chen2023diffusionpcr} uses diffusion to directly regress transformation parameters, while DiffPCR~\cite{diffpcr} employs optimal transport for supervision to enhance correspondence reliability. However, few of these methods generalize well to outdoor scenarios. The challenges primarily arise from the huge point scale of LiDAR points, where establishing the dense correspondences for outdoor LiDAR commonly leads to unaffordable increasing computational costs, hindering the real-world applications.

To address these challenges, we propose a novel probabilistic method named \textbf{E}fficient \textbf{A}utogressive \textbf{D}iffusion for outdoor LiDAR Point Cloud \textbf{Reg}istration (\textbf{EADReg}) , which combines GMM-based outlier correspondence rejection in the coarse layer with the diffusion-based robust point-to-point correspondence estimation in the refinement layers. Specifically, we adopt a coarse-to-fine registration paradigm, downsampling the point cloud pairs through hierarchical networks~\cite{HRegNet}. To perform reliable coarse registration, it's crucial to effectively remove noisy points. Unlike previous methods that introduce extra parameters to identify outliers~\cite{OMNet,UTOPIC,Rethinking_2023}, we design a training-free \textbf{B}i-directional \textbf{G}aussian \textbf{M}ixture \textbf{M}odels Outlier Removal module ({BGMM}) to reject outliers. By filtering the outlier GMM distributions and their associated points, we can perform reliable coarse registration based on the features and geometric information of the remaining points.


Moreover, inspired by diffusion-based temporal motion prediction methods~\cite{Tang_2024,BeLFusion,DiffusionPoser}, we re-think diffusion-based PCR as an autoregressive process. For inference, the diffusion model generates a robust point correspondence matrix between adjacent sampling steps, which are decomposed into translation and rotation components to warp the input source points, forming the source points for the next step. By learning the posterior distribution of ground truth correspondence matrixs, our model iteratively refines the registration, with each transformation preserved in a history buffer. The final result is derived from the sequential product of all transformations. 
Our contributions are as follows:
\begin{itemize}

\item We propose EADReg, a novel outdoor PCR method that integrates Bi-directional Gaussian Mixture Models (BGMMs) to filter out outliers and treats single-frame PCR as a sequential registration process within an autoregressive diffusion paradigm, ensuring strong robustness to perturbations in outdoor LiDAR point clouds.


\item Benefiting from reliable coarse registration, EADReg focuses on the top-K nearest neighbors of the source points to find true correspondences, significantly reducing computational complexity and memory usage. This results in computational speed comparable to convolution-based methods while maintaining high precision.

\item Extensive experiments on the KITTI and NuScenes benchmark datasets demonstrate that our method outperforms state-of-the-art approaches, showcasing convincing performance in challenging outdoor scenarios.

\end{itemize}

\section{Related Work}

PCR is a fundamental research task in 3D computer vision field. In this section, we respectively discuss prior \emph{Learning-based PCR Methods}, \emph{GMMs-based PCR Methods}, and \emph{Diffusion-based PCR Methods}.

\textbf{Learning-based PCR Methods.}\; 
Traditional methods like Iterative Closest Point (ICP)~\cite{ICP} and RANSAC~\cite{RANSAC} are designed for the registration task. However, these methods require either favorable initial transformations or point-level correspondences.
With the advent of deep learning techniques, learning-based methods~\cite{GeoTransformer,Lepard,HRegNet,liu2023contrastive} have achieved significant success in the PCR task. Relying on powerful backbones like PointNet~\cite{Pointnet}, KPConv~\cite{KPConv}, and Point TransFormer~\cite{Pointtransformer}, these methods generally downsample the original point cloud into superpoints and generate associated features with informative information. Subsequently, the transformation can be derived by constructing accurate point-to-point (P2P) correspondences.
Unlike indoor and object-level registration, outdoor point clouds scanned from LiDAR sensors typically exhibit higher sparsity and larger point magnitude, leading to more challenging registration issues. To address these challenges, specific designs have been made to ease the difficulty and reduce the network burden during training. HRegNet~\cite{HRegNet} and RegFormer~\cite{Regformer1} introduce a coarse-to-fine framework to perform registration hierarchically, which accommodates the training process with increasing point magnitudes. 
However, employing prevalent powerful Transformer architecture to extract representative features is inevitable leading to slow convergence and significant memory usage.

\textbf{GMMs-based PCR Methods.}\;
Unlike point-level registration, GMMs-based PCR methods model point clouds via the Expectation-Maximization (EM) optimization algorithm into several clusters and perform alignment based on the modeling results. DeepGMR~\cite{Deepgmr} estimates transformations by minimizing the divergence between GMMs of source and target point clouds. JRMPC~\cite{JRMPC} and FilterReg~\cite{Filterreg} fit point clouds to GMM distributions using Maximum Likelihood Estimation (MLE).
However, due to the sparsity of outdoor LiDAR point cloud scenes, the characteristics of different distributions can be inconspicuous,
leading to degraded registration results on outdoor datasets. G3Reg~\cite{G3reg} introduces a variant of GMMs named Gaussian Ellipsoid Model with Pyramid framework to improve the outdoor LiDAR performance.
Considering the drawbacks of GMMs, directly utilizing GMMs for outdoor LiDAR registration can be challenging. Hence, we take advantage of the modeling ability of GMMs and reject the points contained in the outlier GMMs. Eliminating the misleading effect caused by the outliers, the coarse registration can be easily conducted with the source and target points.

\textbf{Diffusion-based PCR Methods.}\; 
The generative model DDPMs~\cite{ddpm} has made
great development in many fields, including image and semantic matching~\cite{diffmatch,hedlin2024unsupervised,luo2024diffusion}, human motion
estimation~\cite{Tang_2024,BeLFusion,DiffusionPoser}, camera pose estimation~\cite{posediff}, etc.
Recently, A few attempts~\cite{diffpcr,she2024pointdifformer,se3diff} have been made to extend diffusion into point cloud registration (PCR) for accurate dense correspondence generation and translation matrix regression. DiffusionPCR~\cite{chen2023diffusionpcr} introduces diffusion as a reliable regressor which directly generates the transformation parameters through the extracted representative point features. SE3Diff~\cite{se3diff} demonstrate that the accurate diffusion process towards transformation matrices should be strictly constrained by the Lie algebra regulation. DiffPCR~\cite{diffpcr} utilizes OT to construct the global P2P correspondence ground truth as supervision and forces the diffusion to generate the reliable correspondence. These methods show that DDPMs are capable of dealing with the instance-level PCR tasks.
However due to the sparsity and low discriminability of LiDAR point cloud characteristics, 
these methods can hardly generalize towards outdoor LiDAR scenarios. 
Autoregressive diffusion (AR)~\cite{hoogeboom2021autoregressive,AMD,DiffusionPoser,wu2023ar} presents fascinating characteristics topological nodes and sequential motion generation, 
Inspired by this observation, EADReg converts the single frame PCR into a sequence modeling task and achieves superior registration performance through AR's inference paradigm in the outdoor scenarios.

\begin{figure*}
\centering
\includegraphics[scale=0.57]{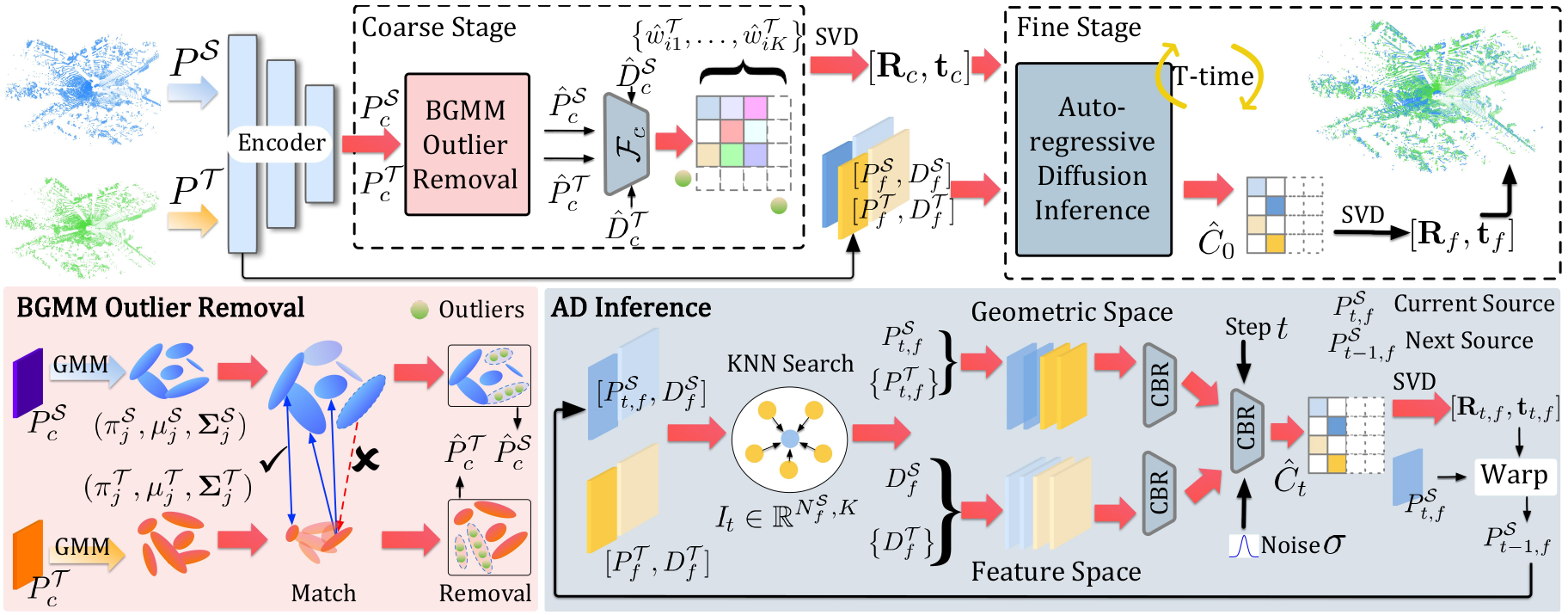}
\caption{The detector-descriptor backbone hierarchically downsamples the input point cloud pairs and extracts corresponding features. Next, the BGMM Outlier Removal module purifies the input points and leverages the predictor network $\mathcal{F}_{c}$ along with SVD to perform coarse registration. Finally, our proposed AD inference framework autoregressively generates regional correspondences and predicts the final registration result.}
\label{fig.wholepipe}

\end{figure*}

\section{EADReg}
\label{sec:method}
\subsection{Network Pipeline} 
Establishing dense point-to-point correspondence is infeasible and computationally-denied due to the huge point scales in outdoor scenes. Therefore, EADReg adopts a coarse-to-fine paradigm, which firstly utilizes the descriptor-detector backbone~\cite{HRegNet} to hierarchically down-sample the point clouds and extract the corresponding multi-scalefeatures. 
Specifically, given source and target point clouds $P^{\mathcal{S}},P^{\mathcal{T}}\in\mathbb{R}^{N\times 3}$, 
the detector utilizes Weighted Furthest Point Sampling (WFPS)~\cite{detector} to downsample the point clouds to different scales of superpoints $P_{l}\in \mathbb{R}^{N_{l}\times 3}$, then the descriptor module extracts their corresponding descriptor $D_{l}\in \mathbb{R}^{N_{l}\times C}$ and uncertainty values $U_{l}\in \mathbb{R}^{N_{l}}$, where $ N_{l}, C_{l}$ are the number of superpoint and channel, the corner mark $l \in \left\{c,f\right\}$ indicates the coarse and fine registration stage.
The details of the detector-descriptor backbone will be provided in supplementary material. 

Since the PCR task has a demand for high inference speed, the prediction networks ${\mathcal{F}}_{c}$ we used in the coarse and ${\mathcal{F}}_{f}$ in fine stage are similar and only constitute with several light-weight CBR (Convolution, Batch Normalization and ReLU) modules, the details of ${\mathcal{F}}_{f}$ is shown in the bottom of Fig.~\ref{fig.wholepipe}, the forward process will be discussed in the Sec.~\ref{sec.gmmblock} and Sec.~\ref{sec.Diff}. 

\subsection{Bi-directional GMMs for Outlier Removal} 
\label{sec.gmmblock}

In previous works, GMMs are widely used to perform registration between point cloud pairs~\cite{Deepgmr,JRMPC,G3reg,Filterreg}. 
Specifically, they first model the point cloud pairs with GMMs, then use the Expectation Maximization (EM) algorithm~\cite{EM} to estimate the real transformation between GMM clusters, which alternates between Expectation step and Maximization step for several iterations. 

However, the estimation performance of the GMMs-based registration method heavily relies on the modeling quality of GMMs. 
Due to the presence of noise and outliers in outdoor scenes,
performing registration between the GMMs group of the LiDAR point cloud pairs will lead to the poor performance. 
In this work, instead of leveraging GMMs as a predictor to estimate the transformation matrix, we adopt it as an outlier-rejection module based on the geometric features of GMMs. Unlike previous learning-based outlier prediction methods~\cite{OMNet,UTOPIC,Rethinking_2023}, we introduce no extra parameters leading to high efficiency.
Specifically, GMMs are first used to establish a multi-modal probability distribution over 3D space which can be represented as a weighted sum of $J$ Gaussian distributions as follows:
\begin{equation}
    p(\mathbf{x} \mid \Theta) := \sum_{j=1}^J \pi_j \mathcal{N}(\mathbf{x} \mid \mathbf{\mu}_j, \mathbf{\Sigma}_j),
\end{equation}
where $\mathbf{x},\Theta$ represent the points and GMMs parameters respectively. $\Theta$ consists of $J$ triplets $(\pi_j,\mathbf{\mu}_j,\mathbf{\Sigma}_j)$, where $\pi_j$ is a scalar mixture weight and $\sum_j \pi_j = 1$, $\mathbf{\mu}_j\in\mathbb{R}^{3\times1}$ is the mean vector and $\mathbf{\Sigma}_j\in\mathbb{R}^{3\times3}$ is the covariance matrix of the $j$-th component.

For filtering out the outliers,
simply using bi-directional first nearest nei maching can lead to plenty of misjudgements,
due to the randomness of the probabilistic model.
We loose the strict rules by only removing the GMMs which is not the top-K neighbor of its nearest counterpart as:
\begin{equation}
\mathcal{G}_{out}^\mathcal{S} = \left\{ 
\Theta_{i}^{\mathcal{S}} \in \Theta^{\mathcal{S}} \mid 
\Theta_{i}^{\mathcal{S}} \notin N_k(\Theta_{j}^{\mathcal{T}}), \Theta_{j}^{\mathcal{T}} = N_1(\Theta_{i}^{\mathcal{S}}) 
\right\}.
\end{equation}
After obtaining the outlier GMMs, we can filter out the attached outlier points  $p_{i,c}^\mathcal{S} \in \mathcal{G}_{out}^\mathcal{S}$, where the purified points are denoted as $\hat{p}_{i,c}^\mathcal{S}$
The outlier GMMs $\mathcal{G}_{out}^\mathcal{T}$ and the purified points $\hat{p}_{i,c}^\mathcal{T}$ are computed in the same way.
Several removal results are presented in Sec.~\ref{sec:Experiments}.

Based on the purified point clouds $\hat{p}_{i,c}^\mathcal{S}, \hat{p}_{i,c}^\mathcal{T}$, we can further predict the truly P2P correspondence for coarse registration. Firstly, We perform KNN search on the feature space $\hat{d}_i^\mathcal{S}$ to retrieve the top-K nearest candidates $\left\{\hat{d}_{i1}^\mathcal{T}, \dots, \hat{d}_{iK}^\mathcal{T}\right\}$, their corresponding geometric points $\left\{\hat{p}_{i1}^\mathcal{T}, \dots, \hat{p}_{iK}^\mathcal{T}\right\}$, and uncertainty value $\left\{\hat{u}_{i1}^\mathcal{T}, \dots, \hat{u}_{iK}^\mathcal{T}\right\}$.
We can firstly construct their geometric features as:
\begin{equation}
\label{eq:geo_fea}
F_{G} =  \left[ \hat{p}_{c}^\mathcal{S}, \left\{\hat{p}_{c}^\mathcal{T}\right\}, p_{c}^\mathcal{S} - \left\{\hat{p}_{c}^\mathcal{T}\right\}, \| p_{c}^\mathcal{S} - \left\{\hat{p}_{c}^\mathcal{T}\right\} \|_2 \right],
\end{equation}
where $\left[ \cdot \right]$ denotes the concatenation operation and $\left\{ \cdot \right\}$ denotes the candidates group.
By introducing the cosine similarity function, we can also measure the similarity between the source features and their top-K nearest neighbors as:
\begin{equation}
S_{ij} = \frac{\langle \hat{d}_i^S, \hat{d}_j^T \rangle}{\| \hat{d}_i^S \|_2 \| \hat{d}_j^T \|_2}, (i \in \hat{N}^{\mathcal{S}}_{c}, j \in iK)
\end{equation}
The descriptor features can be constructed as:
\begin{equation}
\label{eq:fea_fea}
F_{D} =  \left[ \hat{d}_{c}^\mathcal{S}, \left\{\hat{d}_{c}^\mathcal{T}\right\}, \hat{u}_{c}^\mathcal{S} , \left\{\hat{u}_{c}^\mathcal{T}\right\},S \right]
\end{equation}
Then the confidence scores for each candidate is predicted by the prediction network ${\mathcal{F}}_{c}$ as:
\begin{equation}
\left\{\hat{w}_{i1}^\mathcal{T}, \dots, \hat{w}_{iK}^\mathcal{T}\right\}_{i \in \hat{N}_{c}^{\mathcal{S}}}=\mathrm{Softmax}({\mathcal{F}}_{c}[F_{G},F_{D}])
\end{equation}
Where 
$\hat{w}_{ij}^\mathcal{T}$ denotes the similarity of the top-$j$ neighborhood with $\hat{p}_{i,c}^{\mathcal{S}}$.

We can use the weighted sum operation to generate the counterpart of $\hat{p}_{i,c}^{\mathcal{S}}$
\begin{equation}
\label{eq:fusep}
\hat{p}_{i,c}^{fuse} = \sum_{k=1}^{K} \hat{w}_{ik}^\mathcal{T}\hat{p}_{ik}^\mathcal{T}
\end{equation}
We also obtain the fused features $\hat{d}_i^{fuse}$ in the same way and utilize MLP layers to predict the confidence weights for coarse registration
\begin{equation}
\label{eq:confidence}
\hat{w}_{i}=\textbf{MLP}([\hat{d}_i^{fuse}])
\end{equation}


The coarse transformation $\textbf{R}_{c},\textbf{t}_{c}$ can be calculted through the Weighted SVD decompostion:
\begin{equation}
\label{eq:coarseregis}
\textbf{R}_{c}, \textbf{t}_{c} = \min_{\textbf{R}, \textbf{t}} \sum\nolimits  \hat{w}_{i} \lVert \textbf{R} \cdot \hat{p}^{\mathcal{S}}_{i,c} + \textbf{t} - \hat{p}^{fuse}_{i} \rVert_2^2 
\end{equation}

\subsection{Efficient Autoregressive Diffusion} 
\label{sec.Diff}
Previous correspondence-based registration methods~\cite{Cofinet,GeoTransformer,diffpcr} focus on directly generating the global P2P correspondence with dimension $N^{\mathcal{S}}\times N^{\mathcal{T}}$, 
However, the diffusion-based methods are roundly criticized for the low convergence speed, 
predicting global dense P2P correspondences will aggravate the burden of training computation as in the Sec.\ref{sec.abs}, especially for large-scale scenes.
To tackle this problem, we observe that the top-K neighbors searching in source frame are effective enough to find accurate correspondences for each point in target frame, based on the reliable coarse registration results.



\noindent \textbf{Diffusion Correspondence:} 
In this paper, we formulate the correspondence generation as a denoising process $p_{\theta}(C_{t-1}|C_t,\textbf{c})$. Specifically, 
a neural network ${\mathcal{F}}_{f}(C_t,t,F_{G},F_{D})$ is designed by taking the geometric features $F_{G}$ and descriptor features $F_{D}$ as conditions, where the intermediate correspondence is denoted as $C_t \in \mathbb{R}^{N^{\mathcal{S}}_{f}\times K}$. Finally, the output distribution of $\tilde{C}_{0} \in \mathbb{R}^{N^{\mathcal{S}}_{f}\times K}$ is generated progressively through the denoising process. 


Specifically, source points $p_{i,t,f}^\mathcal{S},$ firstly retrieve their top-K nearest candidates$\left\{p_{i1}^\mathcal{T}, \dots, p_{iK}^\mathcal{T}\right\}$, corresponding descriptors $\left\{d_{i1}^\mathcal{T}, \dots, d_{iK}^\mathcal{T}\right\}$ and uncertainty values $\left\{{u}_{i1}^\mathcal{T}, \dots, {u}_{iK}^\mathcal{T}\right\}$.
Similar to the coarse stage, the geometric features can be constructed as: 
\begin{equation}
\label{eq:geo_fea_fine}
F_{G}^t =  \left[ p_{t,f}^\mathcal{S}, \left\{p_{t,f}^\mathcal{T}\right\}, p_{t,f}^\mathcal{S} - \left\{p_{t,f}^\mathcal{T}\right\}, \| p_{t,f}^\mathcal{S} - \left\{p_{t,f}^\mathcal{T}\right\} \|_2 \right].
\end{equation}

Take computational overhead into consideration, in the fine stage, we no longer calculate the similarities, the descriptor features can be constructed as:
\begin{equation}
F_{D} =  \left[d_{f}^\mathcal{S}, \left\{d_{f}^\mathcal{T}\right\}, u_{f}^\mathcal{S} , \left\{u_{f}^\mathcal{T}\right\} \right].
\end{equation}

Finally, by fusing the features with the noisy correspondence $C_t$ and the time $t$, we can get the denoised correspondence $\tilde{C}_{0}$ through ${\mathcal{F}}_{f}[C_t,t,F_{G}^t,F_{D}]$.

\noindent \textbf{Autogressive Inference:}
Inspired by diffusion-based temporal motion prediction methods~\cite{Tang_2024,BeLFusion,DiffusionPoser}, we
rethink the diffusion-based PCR as a autoregressive process. During inference, 
the diffusion model will generate robust point correspondence matrix between adjacent two adjacent sampling steps. The generated correspondences are then decomposed into translation and rotation to warp the input source points, which will constitute the next step's source point.
%
Specifically, obtaining the denoised correspondence from ${\mathcal{F}}_{f}$, 
we can perform the $\mathrm{Softmax}$ operation on the correspondence $\tilde{C}_{0}$ to get the similarity towards top-K candidates of $p_{i,t,f}^\mathcal{S}$ as:
\begin{equation}
\label{eq:softfin}
    \left\{w_{i1}^\mathcal{T}, \dots, w_{iK}^\mathcal{T}\right\}_{i \in N_{f}^{\mathcal{S}}}=\mathrm{Softmax}({\mathcal{F}}_{f}[C_t,t,F_{G}^t,F_{D}]).
    \end{equation}
Similar to Eq.~\ref{eq:fusep}, \ref{eq:confidence} and \ref{eq:coarseregis}, the transformation matrix
$\left\{(\textbf{R}_{t-1,f}, \textbf{t}_{t-1,f}) \mid  t \in T \right\}$
can be derived as: 
\begin{equation}
\label{eq:regisfine}
\textbf{R}_{t-1,f},\textbf{t}_{t-1,f}=\mathop{\arg\min}_{\textbf{R},\textbf{t}} \sum_{i}^{N_{\mathcal{S}}} {w}_{i,t}\left\|\textbf{R}p_{i,t,f}^\mathcal{S}+\textbf{t}-{p}_{i,f}^{fuse}\right\|_2,
\end{equation} 
Where $p_{i,t,f}^\mathcal{S}$ denotes the source point warped by the current transformation as:
\begin{equation}
    p_{i,t,f}^\mathcal{S}=\tilde{\textbf{R}}_{t-1,f}p_{i,f}^{S}+\tilde{\textbf{t}}_{t-1,f}.
\end{equation}
The transformation $\left\{\tilde{\textbf{Tr}}_{t-1}=(\tilde{\textbf{R}}_{t-1},\tilde{\textbf{t}}_{t-1})\right\}$ is built on the previous transformation history as:
\begin{equation}
\tilde{\textbf{Tr}}_{t-1} =  \prod_{i=T}^{t-1} \textbf{Tr}_{i}, \tilde{\textbf{Tr}}_{T}=\textbf{Tr}_{T}=\textbf{Tr}_{c}.
\end{equation}
Since the transformation is derived from the denoised correspondence, we can formulate the autoregressive inference paradigm as:
\begin{equation}
p(C^T, ..., C^t) = \prod_{t} p(C^t \mid C^T, ..., C^{t-1}).
\end{equation}

In this case, the prediction of the correspondence is based on the correspondence history, which fulfills the general form of autoregression. We refer to this inference manner as Autoregressively Inference. 

\subsection{Loss Function}
The overall loss function can be written as: 
\begin{equation}
\label{eq:lossfunc}
    \mathcal{L}=\mathcal{L}_\text{trans}+\alpha \mathcal{L}_\text{rot}+ \mathcal{L}_\text{diff},
\end{equation}
where $\mathcal{L}_\text{trans}$ and $\mathcal{L}_\text{rot}$ are translation loss and rotation loss, respectively. $\mathcal{L}_\text{diff}$ is the loss in diffusion model training. Specifically, given each estimated transformation in the forward process $\tilde{\mathbf{R}},\tilde{\mathbf{t}}$ and the ground truth $\mathbf{R},\mathbf{t}$, $\mathcal{L}_\text{trans}$ and $\mathcal{L}_\text{rot}$ can be calculated as:
\begin{align}
    \mathcal{L}_\text{trans} &= \left\|\mathbf{t}-\tilde{\mathbf{t}}_{l}\right\|_2, l\in \left\{c,f\right\},
\label{eq:ltrans}
    \\
    \mathcal{L}_\text{rot} &= \left\|\tilde{\mathbf{R}}^T\mathbf{R}_l-\mathbf{I}\right\|_2, l\in \left\{c,f\right\},
    \label{eq:lrot}
\end{align}
where $\mathbf{I}$ denotes identity matrix. 
The training loss for the diffusion model $\boldsymbol{\mu_{f}}$ is presented as: 
\begin{equation}
\mathcal{L}_\text{diff} = \mathbb{E}_{t \sim [0, T]} \| {\mathcal{F}}_{f}[C_t,t,F_{G}^t,F_{D}] - C_{gt} \|^2.
\end{equation}
In order to get the supervision $C_{gt}$, we  first use the GT transformation $\mathbf{R},\mathbf{t}$ to warp the source superpoints as $\mathit{P}^{S}_{gt}=\mathbf{R}\mathit{P}^{S}+\mathbf{t}$. Then, we obtain the global correspondence matrix $\tilde{C}_{gt}$ by utilizing the Optimal Transport algorithm~\cite{Optimaltransport} to refine the distance matrix between $\mathit{P}^{S}_{gt}$ and $\mathit{P}^{T}$. Finally, each source superpoint $\mathit{P}^{S}_{gt}$ searches its $K$ nearest target superpoints $\mathit{P}^{T}$ to build the local GT corresponding $C_{gt} \in \mathbb{R}^{N^{\mathcal{S}}_{f}\times K}$.

\begin{algorithm}[t]
\caption{EADReg: Autogregressive Inference}
\textbf{Require:} Registration $\textbf{R}_{c}, \textbf{t}_{c}$ from the coarse stage; Point clouds $P_{f}^{\mathcal{S}} \in \mathbb{R}^{N_{f}^{\mathcal{S}}\times 3}, P_{f}^{\mathcal{T}} \in \mathbb{R}^{N_{f}^{\mathcal{T}}\times 3}$ and corresponding descriptors $D_{f}^{\mathcal{S}}, D_{f}^{\mathcal{T}}$ with uncertainty values $U_{f}^{\mathcal{S}}, U_{f}^{\mathcal{T}}$.\\ 
\textbf{Target:} the denoised correspondence $\tilde{C}_0 \in \mathbb{R}^{N^{\mathcal{S}}_{f}\times K}$.
\begin{algorithmic}[1]
\State $C_T \sim \mathcal{N}(0, 1)^{N \times M}$ 
\State $P_{T,f}^{\mathcal{S}}=\textbf{R}_{c}P^{\mathcal{S}}_{f}+\textbf{t}_{c} \rightarrow F_{G}^T  $ 
\State $F_{D} =  \left[D_{f}^\mathcal{S}, \left\{D_{f}^\mathcal{T}\right\}, U_{f}^\mathcal{S} , \left\{U_{f}^\mathcal{T}\right\} \right]$
\For{$t = T, \dots, 1$}
    \State $\hat{C}_0 \leftarrow {\mathcal{F}}_{f}[C_t,t,F_{G}^t,F_{D}]$
        \State $\textbf{Tr}_{t-1} \leftarrow \textbf{R}_{t-1,f},\textbf{t}_{t-1,f} \leftarrow \hat{C}_0$ \Comment{Equation (\ref{eq:fusep},\ref{eq:confidence},\ref{eq:softfin},\ref{eq:regisfine})}
    \State $\tilde{\textbf{Tr}}_{t-1} =  \prod_{i=T}^{t-1} \textbf{Tr}_{i}, \tilde{\textbf{Tr}}_{0}=\textbf{Tr}_{0}=\textbf{Tr}_{c} $
    \State $    P_{t-1,f}^\mathcal{S}=\tilde{\textbf{R}}_{t-1,f}P_{f}^{S}+\tilde{\textbf{t}}_{t-1,f} \rightarrow F_{G}^{t-1}
$
    \State $\epsilon_t \leftarrow \frac{\hat{C}_0 - \sqrt{\bar{\alpha}_t} \hat{C}_t}{\sqrt{1 - \bar{\alpha}_t}}$
    \State $\hat{C}_{t-1} \leftarrow \sqrt{\alpha_{t-1}} \hat{C}_0 + \sqrt{1 - \alpha_{t-1} - \sigma_t^2} \epsilon_t + \sigma_t z_t$

\EndFor
\end{algorithmic}
\end{algorithm}

\begin{table*}[ht]
\centering
	\caption{Registration performance on KITTI dataset and NuScenes dataset.}
	\label{tab:mainregis}
 \begin{adjustbox}{width=\textwidth}

	\begin{tabular}{r|cccc|cccc}
\toprule



\multirow{2}{*}{Methods} & \multicolumn{4}{c|}{KITTI dataset} & \multicolumn{4}{c}{NuScenes dataset} \\

		\cmidrule(r){2-5}
		\cmidrule(r){6-9}

~ & RTE (m) $\downarrow$ & RRE (deg)$\downarrow$ & Recall $\uparrow$& Time (ms) $\downarrow$& RTE (m) $\downarrow$& RRE (deg) $\downarrow$& Recall$\uparrow$ & Time (ms) $\downarrow$\\
\midrule

		ICP\cite{ICP}  & $0.04\pm 0.05$ & $0.11\pm 0.09$ & 14.3\% & 472.2 & $0.25\pm 0.51$ & $0.25\pm 0.50$ & 18.8\% & 82.0 \\
		FGR\cite{FGR}  & $0.93\pm 0.59$ & $0.96\pm 0.81$ & 39.4\% & 506.1 & $0.71\pm 0.62$ & $1.01\pm 0.92$ & 32.2\% & 284.6 \\
		RANSAC\cite{RANSAC} & $0.13\pm 0.07$ & $0.54\pm 0.40$ & 91.9\% & 549.6 & $0.21\pm 0.19$ & $0.74\pm 0.70$ & 60.9\% & 268.2 \\
		
		\midrule
		DCP\cite{dcp}  & $1.03\pm 0.51$ & $2.07\pm 1.19$ & 47.3\% & 46.4 & $1.09\pm 0.49$ & $2.07\pm 1.14$ & 58.6\% & 45.5\\
		IDAM\cite{IDAM}  & $0.66\pm 0.48$ & $1.06\pm 0.94$ & 70.9\% & 33.4 & $0.47\pm 0.41$ & $0.79\pm 0.78$ & 88.0\% & 32.6 \\
		FMR\cite{FMR}  & $0.66\pm 0.42$ & $1.49\pm 0.85$ & 90.6\% & 85.5 & $0.60\pm 0.39$ & $1.61\pm 0.97$ & 92.1\% & 61.1 \\ 
		DGR\cite{DGR}  & $0.32\pm 0.32$ & $0.37\pm 0.30$ & 98.7\% & 1496.6 & $0.21\pm 0.18$ & $0.48\pm 0.43$ & 98.4\% & 523.0\\

RegFormer\cite{Regformer1} & 0.08 $\pm$ 0.11 & 0.23 $\pm$ 0.21 & 99.8\% & 98.3 & 0.20 $\pm$ $--$ & 0.22 $\pm$ $--$ & 99.9\% & 85.6 \\

HRegNet\cite{HRegNet} & 0.047 $\pm$ 0.037 & 0.147 $\pm$ 0.120 & 100\% & 136.0 & 0.110 $\pm$ 0.096 & 0.285 $\pm$ 0.209 & 100\% & 120.1 \\

HDMNet\cite{HDMNet} & 0.050 $\pm$ 0.057 & 0.159 $\pm$ 0.152 & 99.85\% & 120.2 & 0.114 $\pm$ 0.102 & 0.274 $\pm$ 0.206 & 100\% & 102.9 \\

FlyCore\cite{FlyCore} & 0.05 $\pm$ $--$ & 0.28 $\pm$ $--$ & 99.7\% & 154 & 0.17 $\pm$ $--$ & 0.32 $\pm$ $--$ & 99.7\% & 128 \\

		\midrule
		\rowcolor{gray!30}EADReg & $\mathbf{0.040\pm 0.035}$ & $\mathbf{0.116\pm 0.096}$ &$\mathbf{100\%}$&129.4 &  $\mathbf{0.090\pm 0.072}$ & $\mathbf{0.248\pm 0.187}$ & $\mathbf{100\%}$ &112.4\\
\bottomrule
\end{tabular}
\end{adjustbox}
\end{table*}

\section{Experiments}
\label{sec:Experiments}
\subsection{Datasets and Implementation Details}
\textbf{Datasets.}  We conduct extensive experiments on three large-scale point cloud datasets, KITTI~\cite{kitti}, NuScenes~\cite{nuscenes} and Apollo~\cite{apollo}. KITTI odometry dataset consists of 11 sequences (00-10) with ground truth vehicles poses. It is worth noting that, in the outdoor PCR field, the point cloud pairs we used for training, validation and testing are point clouds between a certain interval frame. The interval is set to 10 as in ~\cite{Regformer1,HRegNet}. Additionally, the metrics of the indoor methods like Geotransformer~\cite{GeoTransformer} is different from ours, for fair comparison we only consider outdoor PCR methods.


According to the settings of the KITTI dataset in ~\cite{Regformer1,HRegNet}, sequences 00-05 are used for training, sequences 06-07 for validation and sequences 08-10 for testing. NuScenes dataset comprises 1000 scenes, and we use the first 700 scenes for training, the following 150 scenes for validation, and the last 150 for testing. The Apollo-SouthBay dataset contains six routes, 
where we choose five of them except SunnyvaleBigloop route, to conduct fair experimental comparison following \cite{HRegNet}.

\noindent\textbf{Implementation Details}
We first utilize voxelization for downsampling, the voxel size is set to 0.3m. After that, we randomly sample 16384 points in the KITTI and Apollo datasets and 8192 points in the NuScenes dataset for registration. Our testing procedure for all experiments utilizes DDIM~\cite{DDIM} for acceleration, and the influence of different sampling steps will be discussed in the Sec.\ref{abs}.
The learning rate is initially set to 0.001 with a weight decay of 50\% every 10 epochs. The training framewrok is implemented with PyTorch 2.0.1 and Adam optimizer.  
For training precedure, we first utilize chamfer~\cite{usip} and match loss~\cite{rskdd} to pretrain our detector-descriptor backbone for robust feature extraction, then train the entire model with batch size 16 for 50 epochs iterations on a single node with an NVIDIA GeForce 3090 GPU with Intel Xeon W-2265 CPU.

\begin{figure*}
\centering
\includegraphics[width=0.8\textwidth]{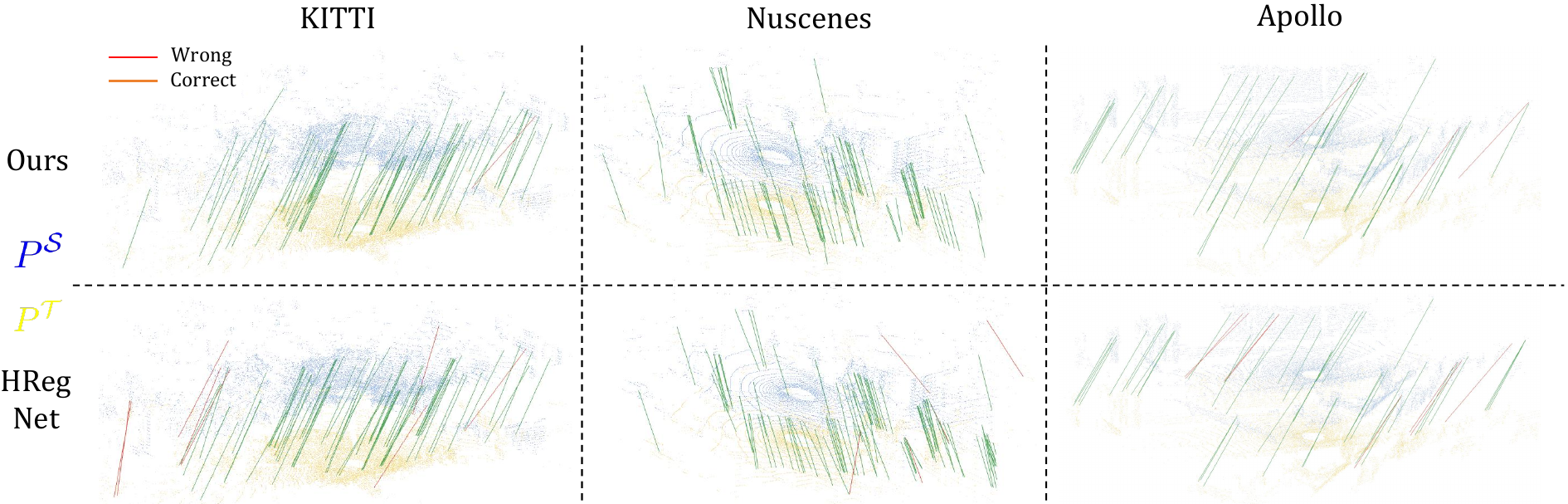}
\caption{Qualitative visualization of registration performance. From left to right, we compare our proposed method with HRegNet using three samples from the KITTI, NuScenes, and Apollo-Southbay datasets, respectively. Specifically, we select only the correspondences with confidence weights $\hat{w}$ greater than 0.001.}
\label{fig.correspondence}

\end{figure*}

\subsection{Evaluation}
\textbf{Qualitative Visualization}
We visualize the registration results with 3 samples from KITTI, Nuscenes and Apollo datasets in Fig.\ref{fig.correspondence}. 

Specifically, we first select the correspondences with confidence weights 
$\hat{w}$ greater than 0.001. Then, we warp the selected source points using the ground truth transformation. If the distance between a warped source point and the target point exceeds 5 meters, the correspondence is considered a wrong prediction; otherwise, it is regarded as correct.

\begin{figure}
\centering
\includegraphics[scale=0.56]{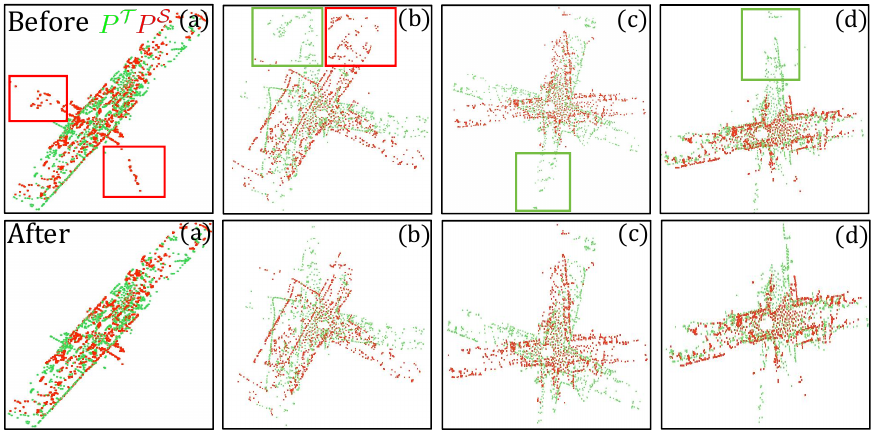}
\caption{We select four samples from the KITTI dataset to illustrate the outlier removal process using the BGMM module. The first row contains the original point clouds, and the second row shows the purified version. The outliers are framed within boxes.}
\label{fig.gmmrevoval}
\end{figure}

To verify the effectiveness of our proposed BGMM Outlier Removal module, we present four removal results in Fig. \ref{fig.gmmrevoval}. From the visualization, it is evident that BGMM successfully removes outlier points and retains all the points relevant for registration.

\textbf{Quantitative Evaluation}
We adopt relative translation
error (RTE), relative rotation error (RRE), registration recal (RR) and average running time on the test point cloud pairs to evaluate the registration performance. 
Specifically, since the failed registrations can result in exceptionally large RRE and RTE values, leading to unreliable error metrics~\cite{HRegNet}. We define successful registration as occurring when both the Relative Translation Error (RTE) and the Relative Rotation Error (RRE) fall within the thresholds $RTE < 2\text{m}$ and $RRE < 5^{\circ}$, respectively.
The registration performance of different methods are presented in Tab.\ref{tab:mainregis}. 


\begin{table}[t]
\centering
\caption{Registration Performance on Apollo-Southbay 
Dataset}
            	\label{tab:apollo}

\resizebox{\columnwidth}{!}{%

\begin{tabular}{r|cccc}

\toprule
                  		Model  & RTE (m) $\downarrow$ & RRE (deg)$\downarrow$ & Recall $\uparrow$& Time (ms) $\downarrow$ \\

\midrule
ICP\cite{ICP}   &$ 0.039 \pm 0.170 $&$ 0.046 \pm 0.257 $&$ 46.45\% $&$ 470.2 $\\
RANSAC\cite{RANSAC}  &$ 0.125 \pm 0.114 $&$ 0.361 \pm 0.308 $&$ 83.72\% $&$ 552.1 $\\
DCP\cite{dcp}  &$ 1.174 \pm 0.499 $&$ 2.155 \pm 1.254 $&$ 28.49\% $&$ 41.0 $\\
IDAM\cite{IDAM}  &$ 0.456 \pm 0.116 $&$ 0.127 \pm 0.429 $&$ 24.03\% $&$ 32.0 $\\
FMR\cite{FMR}  &$ 0.653 \pm 0.188 $&$ 0.727 \pm 0.658 $&$ 83.76\% $&$ 83.4 $\\
DGR\cite{DGR}  &$ 0.132 \pm 0.151 $&$ 0.127 \pm 0.146 $&$ 99.64\% $&$ 257.0 $\\
HRegNet\cite{HRegNet} &$ 0.034 \pm 0.037 $&$ 0.079 \pm 0.079 $&$ 99.88\% $&$ 129.5 $\\
		\midrule

\rowcolor{gray!30}EADReg &$ \mathbf{0.028 \pm 0.031 }$&$ \mathbf{0.064 \pm 0.062} $&$ \mathbf{99.92\%} $&$ 121.6$ \\
\bottomrule
\end{tabular}}
\end{table}

\noindent \textbf{Comparision with baseline methods:}
We choose ICP~\cite{ICP}, RANSAC~\cite{RANSAC} and Fast Global Registration (FGR)~\cite{FGR} as on behalf of
the classical PCR methods, as for learning-based methods, we select DCP~\cite{dcp}, IDAM~\cite{IDAM}, FMR~\cite{FMR}, DGR~\cite{DGR}, RegFormer~\cite{Regformer1}, HRegNet~\cite{HRegNet}, HDMNet~\cite{HDMNet} and Flycore~\cite{FlyCore} for comparision.

According to the results in Tab.\ref{tab:mainregis}, classical methods either achieve poor recall performance on KITTI and Nuscenes datasets (with 14.25\% and 39.42\% for ICP and FGR on KITTI respectively), or generate suboptimal transformation compared with Learning-based methods (with average RTE 0.929 and 0.126 for FGR and RANSAC respectively). 
Moreover, since classical methods require iterative optimization for convergence, learning based methods present better efficiency.

With the development of the learning-based PCR methods, the recall presents a significant upward trend,
and start with RegFormer the recall achieve nearly 100\% on both KITTI and Nuscenes datasets under the generally outdoor PCR setting. 
Specifically, 
Although the DCP, IDAM, and FMR exhibit significantly lower inference times compared to EADReg, the EADReg shows notably higher average RTE and RRE values.
RegFormer, HRegNet, HDMNet and Flycore both achieve centimetre-level RTE performance on KITTI dataset, with 0.08, 0.047, 0.05 and 0.05 respectively.
Our proposed EADReg outperforms all the baselines and achieve 0.040 average RTE and 0.117 RRE degree, exceeding HRegNet by 0.7cm and 0.03 degree. 

Meanwhile, on Nuscenes dataset, EADReg is the only one reaches centimetre-level RTE performance, exceeding HRegNet and RegFormer by 2cm and 11cm respectively. Meanwhile, the variance of EADReg is the smallest across KITTI and Nuscenes, with 3.5cm and 7.2cm respectively.

Additionally, EADReg does not suffer from the slow inference problem commonly associated with diffusion-based models. In fact, the running time of EADReg is faster than FlyCore and HRegNet by 22.8 ms and 4.8 ms, respectively, effectively demonstrating the efficiency of EADReg.

\noindent \textbf{Comparision on Apollo-Southbay dataset:}
Apollo-Southbay dataset has 35109 test data samples and is about six times bigger than KITTI and Nuscenes. Therefore, the results on the Apollo dataset can more accurately reflect the model's performance.

Refer to Tab.\ref{tab:apollo}, EADReg achieves the best performance compared with the baseline methods, surpasses the suboptimal method HRegNet on the average RTE, average RRE and recall with 0.6cm, 0.015 degree and 0.04\% respectively, the variants of EADReg on RTE and RRE are also the smallest, with 3.1cm and 0.06 degree.

Based on comprehensive experiments, it is evident that the newly introduced EADReg surpasses existing methods in terms of accuracy and successfully strikes a favorable balance between performance and efficiency.



\subsection{Ablation Study}
\label{sec.abs}
We perform abundant ablation studies on KITTI dataset to demonstrate the effectiveness of the hierarchical structure and the introduction of the similarity features.

\noindent \textbf{Network Structure:}
We use the output transformation $\mathbf{R},\mathbf{t}$ from the coarse stage and the fine stage respectively for evaluation. 
The registration results
are shown in Tab.\ref{tab:abc2f}. 
Since the average RRE of the coarse stage with 0.119cm is better than the learning based method DGR~\cite{DGR} with 0.32, we can demonstrates that the coarse stage do provide a reliable initial transformation.
\begin{table}[t]
	\centering
	\caption{Ablation studies of the architecture.}
	\label{tab:abc2f}
	\footnotesize
 \resizebox{\columnwidth}{!}{%

	\begin{tabular}{r|cccc}
		\toprule
  		Model & RTE (m) & RRE (deg) & Recall & Time (ms) \\

		\midrule
	  Only Coarse& $0.119\pm 0.102$ & $0.373 \pm 0.284$ &$ 99.97\%$ & $95.5$ \\
		EADReg w/o $GMM$ & $0.042\pm 0.035$ & $0.143\pm 0.118$ &$ 100\%$ & $113.5$ \\
		\midrule
		\rowcolor{gray!30}EADReg& $\mathbf{0.040\pm 0.034 }$& $\mathbf{0.117 \pm 0.107}$ & $\mathbf{100.00\%} $& $129.4$ \\
		\bottomrule
	\end{tabular}}
\end{table}
\begin{table}[t]
	\centering
	\caption{Ablation study results of EADReg on the GMM outlier rejection with different number of clusters $J$. The best results for each criterion are labeled in bold.}
	\label{tab:abgmm}
	\footnotesize
 \resizebox{\columnwidth}{!}{%

	\begin{tabular}{c|cccc}
		\toprule
		Clusters & RTE (m) & RRE (deg) & Recall & Time (ms) \\

		\midrule
        $J =8$  &$ 0.040\pm  0.034$&$\mathbf{0.116\pm0.096}$ &$ 100.00\% $ &$\mathbf{129.4}$\\
        $J =12$  & $0.041\pm 0.035$& $0.117\pm0.094$& $100.00\% $ &$136.2$\\
        $J =16$  & $\mathbf{0.039\pm  0.036}$&$0.115 \pm 0.101$&$ 100.00\%$ &$140.8$\\
        $J =32$ & $0.042\pm  0.039$& $0.123\pm0.110$& $99.96\% $ &$143.5$\\

		\bottomrule
	\end{tabular}}
\end{table}
\begin{table}[t]
    \centering
    \caption{Ablation study results of the hyperparameter $\alpha$'s influence on the registration results on KITTI dataset. The best results for each criterion are labeled in bold.}
            	\label{tab:abalpha}
	\footnotesize

    \begin{tabular}{c|ccc}
        \toprule
                  		$\alpha$ & RTE (m) & RRE (deg) & Recall \\

        \midrule
        $\alpha =1$  & $0.041 \pm 0.038 $&$0.154\pm 0.122$&$ 99.99\%$ \\
        $\alpha =2$  &$ 0.040 \pm0.036 $&$0.133\pm0.111$& $100.00\%$  \\
        $\alpha =3$  & $0.042 \pm 0.036$ &$0.121\pm0.103$&$ 100.00\% $\\
        $\alpha =4$ &$\mathbf{0.040\pm 0.035}$ & $0.116\pm 0.096$&100\% \\
        $\alpha =5$ &$ 0.041 \pm 0.039$ &$0.116\pm0.100$& $99.99\% $ \\
        $\alpha =6$ & $0.043 \pm 0.038$ &$\mathbf{0.100\pm0.089}$& $99.97\%$  \\
        \bottomrule
    \end{tabular}
\end{table}
\begin{table}[t]
    \centering
    \caption{Ablation study results of EADReg with different scale of correspondence region $K$, Occ is the abbreviation of the GPU memory occupancy during training.}
    	\label{tab:abk}
\resizebox{\columnwidth}{!}{%

    \begin{tabular}{c|cccc>{\columncolor{gray!30}}c}
        \toprule
          		$K$ & RTE (m) & RRE (deg) & Recall & Time (ms) &Occ (MB)\\

        \midrule
        $K = 1$ & $0.042 \pm 0.039$ & $0.122 \pm 0.114$ & $100.100\%$ & $128.7$&19083 \\
        $K = 3$ & $\mathbf{0.040\pm 0.034 }$& $\mathbf{0.117 \pm 0.107}$ & $100.00\% $& $129.4$&20583 \\
        $K = 5$ & $0.041 \pm 0.035$ & $0.116 \pm 0.106$ & $99.99\%$ & $131.4$ &22033\\
        $K = 7$ & $0.041 \pm 0.035$ & $0.122 \pm 0.102$ & $100.100\%$ & $132.2$ &23571\\
        \bottomrule
    \end{tabular}}
\end{table}
\begin{table}[t]
    \centering
    \caption{Ablation study results of EADReg with different sampling steps.}
    	\label{tab:absampling}
\resizebox{\columnwidth}{!}{%

    \begin{tabular}{c|cccc}
        \toprule
          		$K$ & RTE (m) & RRE (deg) & Recall & Time (ms) \\

        \midrule
        $S = 1$ & $0.048 \pm 0.040$ & $0.132 \pm 0.111$ & $99.99\%$ & $115.6$ \\
        $S = 2$ & $0.0418 \pm 0.035$ & $0.118 \pm 0.098$ & $99.99\%$ & $123.2$ \\
        $S = 3$ & $0.040 \pm 0.034$ & $0.116 \pm 0.096$ & $100.00\%$ & $129.4$ \\
        $S = 4$ & $0.040 \pm 0.033$ & $0.115 \pm 0.096$ & $99.99\%$ & $136.2$ \\
        $S = 5$ & $0.040 \pm 0.033$ & $0.115 \pm 0.097$ & $99.99\%$ & $141.2$ \\
        $S = 6$ & $0.040 \pm 0.033$ & $0.114 \pm 0.098$ & $99.99\%$ & $148.8$ \\
        \bottomrule
    \end{tabular}}
\end{table}

\noindent \textbf{Number of GMM Clusters:} 
According to the results in Tab.~\ref{tab:abgmm}, although the results show minimal changes as $J$ increases, the inference time increases steadily, from 129.4 ms to 143.5 ms. Thus, we choose $J=8$ for experiments.

\noindent \textbf{Weight of the $\alpha$:} 
EADReg is trained with combinations of translation loss
$\mathcal{L}_{trans}$, rotation loss $\alpha\mathcal{L}_{rot}$ and diffusion loss $\mathcal{L}_{diff}$. 
We leave the ablation study of different set the weights of $\mathcal{L}_{diff}$ and $\mathcal{L}_{trans}$ to 1 for experiments and leave the ablation study in the supplementary materials.
According to Tab.\ref{tab:abalpha},
as the $\alpha$ increases, the average RRE monotonically decreases from 0.154 to 0.100 and RTE reaches the best when $\alpha=4$.
In order to balance the performance between RTE and RRE metrics, we choose $\alpha=4$ for our experiments, which achieves the best RTE with 4cm and suboptimal RRE with 0.115 degree, and the best recall 100\%.

\noindent \textbf{Number of Correspondence Candidate:} 
In this part, we demonstrate the reason why generating the global P2P correspondence$C \in \mathbb{R}^{N^{\mathcal{S}}_{f}\times N^{\mathcal{T}}}$ can be impractical for deployment.
From Table \ref{tab:abk}, we observe that increasing 
$K$ only slightly affects registration performance, supporting our hypothesis that true correspondences lie in the nearest 
top-K neighbors after reliable coarse registration. However, GPU memory usage rises sharply from 19,083 MB to 23,571 MB, indicating that predicting global P2P correspondences will inevitably lead to prohibitive training overhead.


\noindent \textbf{Number of Sampling Steps:} 
We further
conduct several experiments with
different diffusion step numbers and present
the results in Tab.\ref{tab:absampling}. We observe that the performance stabilize after $S=3$, and shows minor improvements
when $S > 3$. Meanwhile, the inference time of EADReg monotonically increasing, from 115.6ms to 148.8ms, 
in order to balance the accuracy and efficiency, we choose $S=3$
in our method.

\section{Conclusion}
In this paper, we provide an efficient diffusion-based network for large-scale outdoor LiDAR point cloud registration named EADReg. EADReg follows a coarse-to-fine registration paradigm, which leverages detector-descriptor backbone to downsample the original point cloud and extract the corresponding features. In the coarse stage, we propose the BGMM module to reject the outlier points. In the fine stage, we introduce autogressive diffusion inference to generate the reliable P2P correspondence. Extensive experiments show that EADReg achieves a favourable balance between performance and efficiency. 

{
    \small

    \bibliographystyle{ieeenat_fullname}
    \bibliography{PaperForReview}
}




\end{document}